\documentclass[10pt,twocolumn,letterpaper]{article}

\usepackage{cvpr}
\usepackage[T1]{fontenc}
\usepackage[utf8]{inputenc}
\usepackage{times}
\usepackage{graphicx}
\usepackage{amsmath}
\usepackage{amssymb}

\usepackage{tabularx}
\usepackage{booktabs}
\usepackage{multirow}
\usepackage{caption}
\usepackage[dvipsnames]{xcolor}
\usepackage{subcaption}

\newcolumntype{Y}{>{\centering\arraybackslash}X}

\usepackage[breaklinks=true,bookmarks=false]{hyperref}
\usepackage[capitalize]{cleveref}
\cvprfinalcopy % *** Uncomment this line for the final submission

\def\cvprPaperID{17} % *** Enter the CVPR Paper ID here

\ifcvprfinal\fi
\begin{document}

%%%%%%%%% TITLE
\title{Joint Learning from Earth Observation and OpenStreetMap Data\\to Get Faster Better Semantic Maps}

\author{Nicolas Audebert, Bertrand Le Saux\\
ONERA, \textit{The French Aerospace Lab}\\
F-91761 Palaiseau, France\\
{\tt\small (nicolas.audebert,bertrand.le\_saux)@onera.fr}
% \and
% Bertrand Le Saux\\
% ONERA, \textit{The French Aerospace Lab}\\
% F-91761 Palaiseau, France\\
% {\tt\small bertrand.le\_saux@onera.fr}
\and
Sébastien Lefèvre\\
Univ. Bretagne-Sud, UMR 6074, IRISA\\
F-56000 Vannes, France\\
{\tt\small sebastien.lefevre@irisa.fr}
}
%$^{1}$ \quad ONERA, \textit{The French Aerospace Lab}, F-91761 Palaiseau, France (nicolas.audebert,bertrand.le\_saux)@onera.fr\\
%$^{2}$ \quad Univ. Bretagne-Sud, UMR 6074, IRISA, F-56000 Vannes, France - sebastien.lefevre@irisa.fr}

\maketitle
\thispagestyle{empty}

%%%%%%%%% ABSTRACT
\begin{abstract}
In this work, we investigate the use of OpenStreetMap data for semantic labeling of Earth Observation images. Deep neural networks have been used in the past for remote sensing data classification from various sensors, including multispectral, hyperspectral, SAR and LiDAR data. While OpenStreetMap has already been used as ground truth data for training such networks, this abundant data source remains rarely exploited as an input information layer. In this paper, we study different use cases and deep network architectures to leverage  OpenStreetMap data for semantic labeling of aerial and satellite images. Especially, we look into fusion based architectures and coarse-to-fine segmentation to include the OpenStreetMap layer into multispectral-based deep fully convolutional networks. We illustrate how these methods can be successfully used on two public datasets: ISPRS Potsdam and DFC2017. We show that OpenStreetMap data can efficiently be integrated into the vision-based deep learning models and that it significantly improves both the accuracy performance and the convergence speed of the networks.
\end{abstract}

%%%%%%%%% BODY TEXT
\section{Introduction}
Dense labeling of remote sensing data is a common task to generate semantic maps of large areas automatically, either to perform cartography of urban areas or to determine land use covers at a large scale.
Many techniques that originated in the computer vision community for semantic segmentation can be applied on Red-Green-Blue (RGB) remote sensing data to this end and has been successfully used with state-of-the-art results, \eg Convolutional Neural Networks (CNN) for land use classification~\cite{castelluccio_land_2015} or Fully Convolutional Networks (FCN) for semantic labeling of urban areas~\cite{audebert_semantic_2016}.
Yet, remote sensing data is rarely limited to RGB channels. Often, multiple sensors are used to obtain complementary physical information about the observed area, with the final products being co-registered multispectral, LiDAR and SAR data. Several works successfully investigated semantic labeling of remote sensing data using fusion of multiple heterogeneous sensors~\cite{audebert_semantic_2016,paisitkriangkrai_effective_2015,hu_fusionet:_2017} to leverage complementary informations sources.
However, there is still a huge untapped source of information: the public geo-information system (GIS) databases that rely on voluntary crowdsourced annotations. These databases, such as the very popular OpenStreetMap (OSM), are an extremely valuable source of information, that is mostly used as a training ground truth~\cite{mnih_learning_2010} as it contains data layers such as roads, buildings footprints or forest coverage.
But this data can also be seen as an alternative input layer that will provide another information channel compared to the classical sensors. Consequently, we investigate here the following question: how can we use OSM data to improve sensor-based semantic labeling of remote sensing data ?

\section{Related work}
Semantic labeling of aerial and satellite images using deep neural networks has been investigated several times in the literature. Since the early work on road extraction using CNN~\cite{mnih_learning_2010}, many studies investigated deep neural networks for automatic processing of aerial and satellite data. The most recent works using deep learning mostly focus on the use of Fully Convolutional Networks~\cite{long_fully_2015}, an architecture initially designed for semantic segmentation of multimedia images, and later successfully applied to remote sensing data at various resolutions. This type of deep models obtained excellent results on very high resolution datasets~\cite{sherrah_fully_2016}, often combined with multiscale analysis~\cite{audebert_semantic_2016}, graphical model post-processing~\cite{volpi_dense_2017} and boundary prediction~\cite{marmanis_classification_2016}. The same architectures were also successful for building and roads extraction from satellite images~\cite{maggiori_fully_2016,vakalopoulou_building_2015,yuan_automatic_2016} at a significantly lower resolution.

Although most of the works are limited to optical data, \eg RGB or multispectral images, the fusion of heterogeneous data layers has also been investigated in several works. Notably, multimodal processing of very high resolution images has been successfully applied to a combination of deep features and superpixel classification in~\cite{campos-taberner_processing_2016}, while a deep framework for the fusion of video and multispectral data was proposed in~\cite{mou_spatiotemporal_2016}. Fusion of multitemporal data for joint registration and change detection was investigated in \cite{vakalopoulou_simultaneous_2016}. Heterogeneous data fusion was also explored through deep features mixed with hand-crafted features for random forest classification~\cite{paisitkriangkrai_effective_2015}, and later using end-to-end deep networks in~\cite{audebert_semantic_2016} for LiDAR and RGB data, and \cite{hu_fusionet:_2017} for hyperspectral and SAR data.

While all these works investigated data fusion of various sensors, they did not study the inclusion of highly processed, semantically richer data such as OpenStreetMap layers. Indeed, since the launch of OpenStreetMap (OSM) and Google Maps in 2004, map data became widely available and have been largely used within remote sensing applications. First, they can be used as targets for deep learning algorithms, such as in the seminal work of Mnih \cite{mnih_learning_2010,mnih_machine_2013}, since those layers already provide accurate information about the buildings and roads footprints. More recent works in simultaneous registration and classification~\cite{vakalopoulou_simultaneous_2016} and large-scale classification~\cite{maggiori_fully_2016,costea_aerial_2016} also rely on OSM data to perform supervised learning. The generation of OpenStreetMap rasters from optical images with Generative Adversarial Networks has also been investigated~\cite{isola_image--image_2016}, but the authors did not evaluate their method with classification metrics as they were only interested by perceptually coherent image-to-image translation. Second, the map layers can be used as inputs of a processing flow to produce new geo-spatial data. Although the coverage and the quality of the annotations from open GIS vary a lot depending on the users' knowledge and number of contributors, this data may contain relevant information for mapping specific areas and classes. A deterministic framework to create land use and land cover maps from crowdsourced maps such as OSM data was proposed in~\cite{fonte_using_2017}. Machine learning tools (a random forest variant) also allow coupling remote sensing and volunteered geographic information (VGI) to predict natural hazard exposure~\cite{geis_joint_2017} and local climate zones~\cite{danylo_contributing_2016}, while active deep learning helps finding unlabeled objects in OSM~\cite{chen_deepvgi:_2017}. However, to the best of our knowledge, no VGI has been used as an input (opposite of a target) in deep neural networks yet.

\section{Method}
\begin{figure*}[t!]
	\centering
    \begin{subfigure}[t]{0.48\textwidth}
	\includegraphics[width=\textwidth]{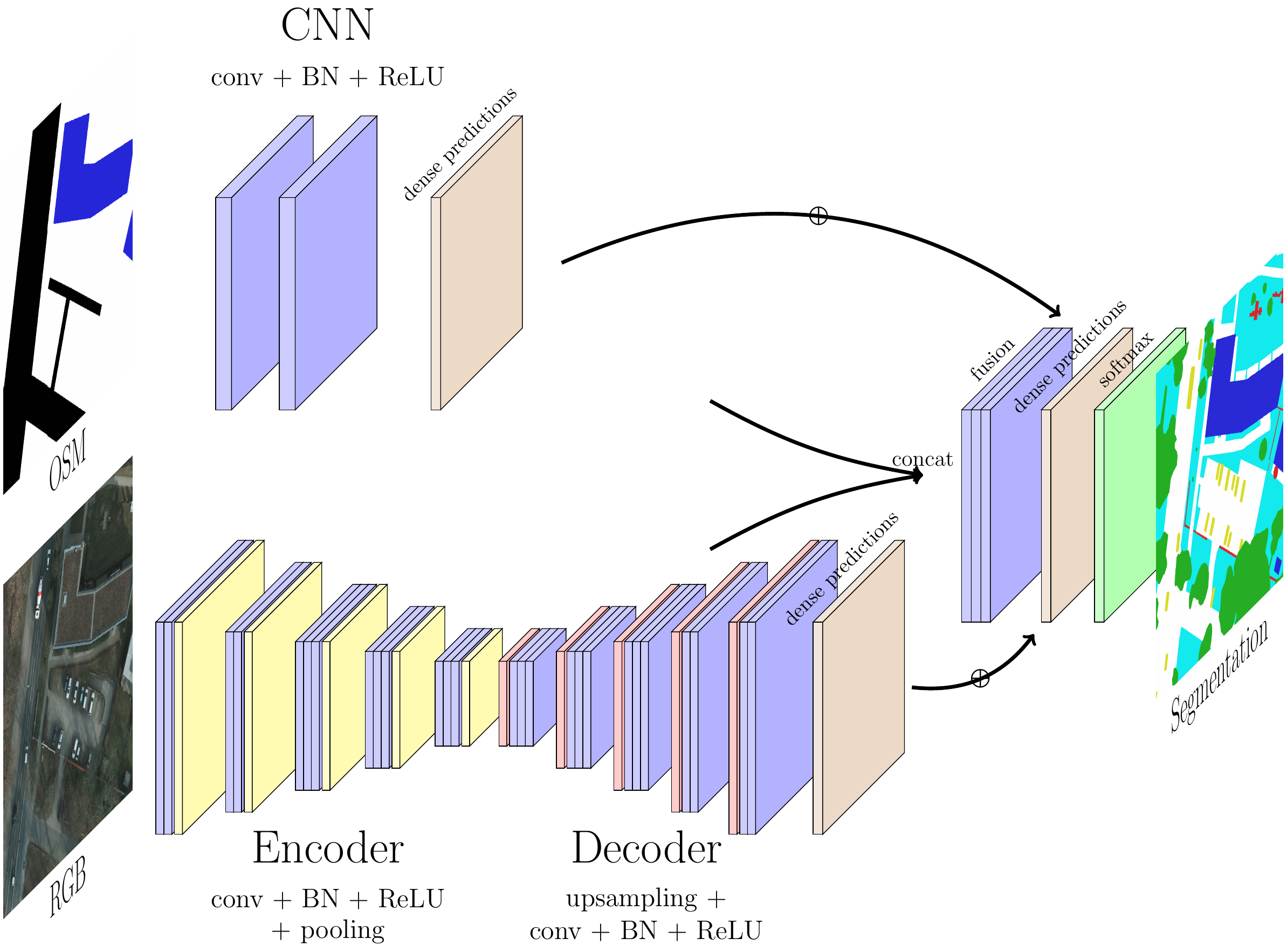}
    \caption{Optical and OSM data fusion using residual correction~\cite{audebert_semantic_2016}.}
    \label{fig:refinet}
    \end{subfigure}
    \hfill
    \begin{subfigure}[t]{0.48\textwidth}
	\includegraphics[width=\textwidth]{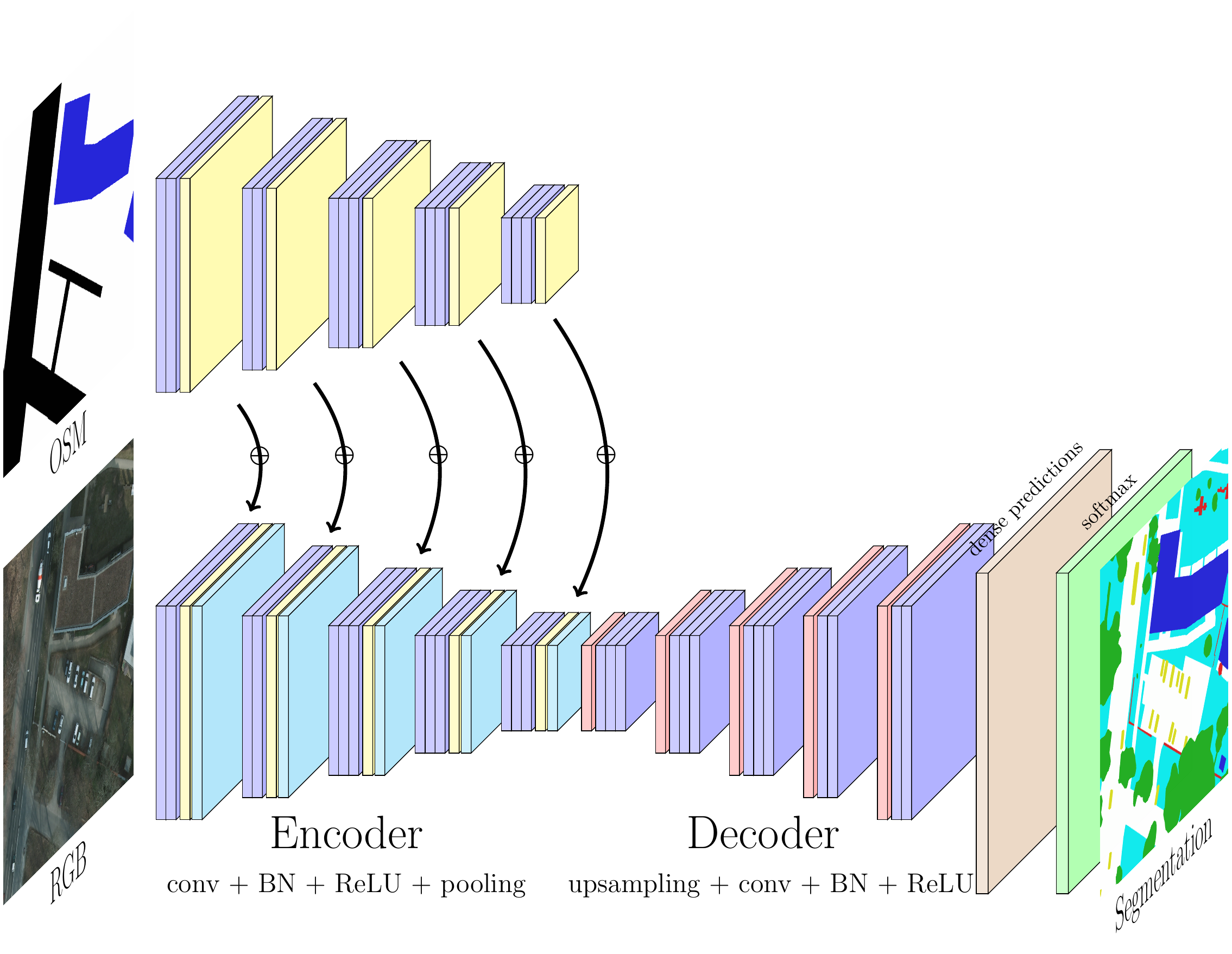}
    \caption{FuseNet~\cite{hazirbas_fusenet:_2016} architecture applied to optical and OSM data.}
    \label{fig:fusenet}
    \end{subfigure}
    \caption{Deep learning architectures for joint processing of optical and OpenStreetMap data.}
\end{figure*}

In this work, we investigate two use cases for highly structured semantic data as input to deep networks. The first one occurs when the semantic data is close to the ground truth, in our case where OpenStreetMap already contains a part of the information we want to label, \eg finding roads in airborne images. The second one occurs when what we want to label can be inferred from OpenStreetMap but only indirectly, for example when we want to derive the type of settlement based on building footprints.

\subsection{Coarse-to-fine segmentation}
In some cases, OpenStreetMap data can be used as a coarse approximation from the ground truth, \eg for objects such as buildings and roads. Therefore, we would need only to learn how to refine this coarse map to obtain the final very high resolution pixel-wise segmentation. Especially, learning only this difference can be seen as a form of residual learning~\cite{he_deep_2016} which should help the training process. This was also suggested in~\cite{lin_refinenet:_2016} to help the classification process by first learning coarse approximations of the ground truth before iteratively refining the predictions.

In our case, we can use a simple FCN with only two layers to convert the raw rasterized OSM data into a semantic map that approximates the ground truth, trained with the usual backpropagation technique. This FCN will be denoted OSMNet in the following sections. OSMNet only manipulates the original OSM map to project it in a representation space compatible with the ground truth classes. The optical data will then be used as input to a FCN model to infer a semantic segmentation that will complete the approximate map derived from OSM. In this work, we will use the SegNet architecture~\cite{badrinarayanan_segnet:_2017} that we will train following the guidelines from~\cite{audebert_semantic_2016}. SegNet is an encoder-decoder deep neural network that projects an input image into a semantic map with the same resolution. Using these two models, we can compute the average prediction computed using the two data sources. In this framework, if we denote $I$ the input image, $O$ the input OSM raster, $P_{opt}$ the prediction function for SegNet and $P_{osm}$ the prediction function for OSMNet, the final prediction $P$ is:
\begin{equation}
P(I, O) = \frac{1}{2} (P_{opt}(I) + P_{osm}(O))~.
\end{equation}

If $P_{osm}(O)$ is already an adequate approximation of the ground truth $GT$, then the training process will try to minimize:
\begin{equation}
P_{opt} \propto GT - P_{osm}(O) \ll GT~,
\end{equation}
which should have a similar effect on the training process than the residual learning from~\cite{he_deep_2016}.

Moreover, to refine even further this average map, we use a residual correction network~\cite{audebert_semantic_2016}. This module consists in a residual three-layer FCN that learns how to correct the average prediction using the complementary information from the OSM and optical data sources. If we denote $C$ the prediction function of the residual correction network, we finally have:
\begin{equation}
P(I, O) = \frac{1}{2} (P_{opt}(I) + P_{osm}(O)) + C(Z_{opt}(I), Z_{osm}(O))~,
\end{equation}
where $Z_{opt}$ and $Z_{osm}$ are the last feature maps from SegNet and OSMNet, respectively.

The residual learning using this module can be seen as learning an error correction to refine and correct occasional errors in the prediction. The full pipeline is illustrated in the~\cref{fig:refinet}. The whole architecture is trained jointly.

\subsection{Dual-stream convolutional neural networks}

FCN with several sources have been investigated several times in the past, notably for processing RGB-D (or 2.5D) images in the computer vision community~\cite{eitel_multimodal_2015}. In this work, we use the FuseNet architecture~\cite{hazirbas_fusenet:_2016} to combine optical and OSM data. It is based on the popular SegNet~\cite{badrinarayanan_segnet:_2017} model. FuseNet has two encoders, one for each source. After each convolutional block, the activations maps from the second encoder are summed into the activations maps from the first encoder. This allows the two encoders to learn a joint representation of the two modalities. Then, a single decoder performs both the spatial upsampling and the pixel-wise classification. As detailed in~\cref{fig:fusenet}, one main branch learns this joint summed representation while the ancillary branch learns only OSM-dependent activations. If we denote $P$ the prediction function from FuseNet, $I$ the input image, $O$ the input OSM rasters, $E_i^{\{opt,osm\}}$ the encoded feature maps after the $i^{th}$ block, $B_i^{\{opt,osm\}}$ the encoding functions for the $i^{th}$ block and $D$ the decoding function:
\begin{equation}
P(I,O) = D(E_5^{opt}(I,O))
%P(I,O) = D(E(I,O))
\end{equation}
and
\begin{equation}
E_{i+1}^{opt}(I,O) = B_i^{opt}(E_i^{opt}(I, O)) + B_i^{osm}(E_i^{osm}(O))~.
\end{equation}
This architecture allows us to fuse both data streams in the internal representation learnt by the network. This model is illustrated in~\cref{fig:fusenet}.

\section{Experiments}

\subsection{Datasets}

\begin{figure*}[t]
	\begin{subfigure}{0.49\textwidth}
    	\begin{subfigure}{0.49\textwidth}
        \includegraphics[width=\textwidth]{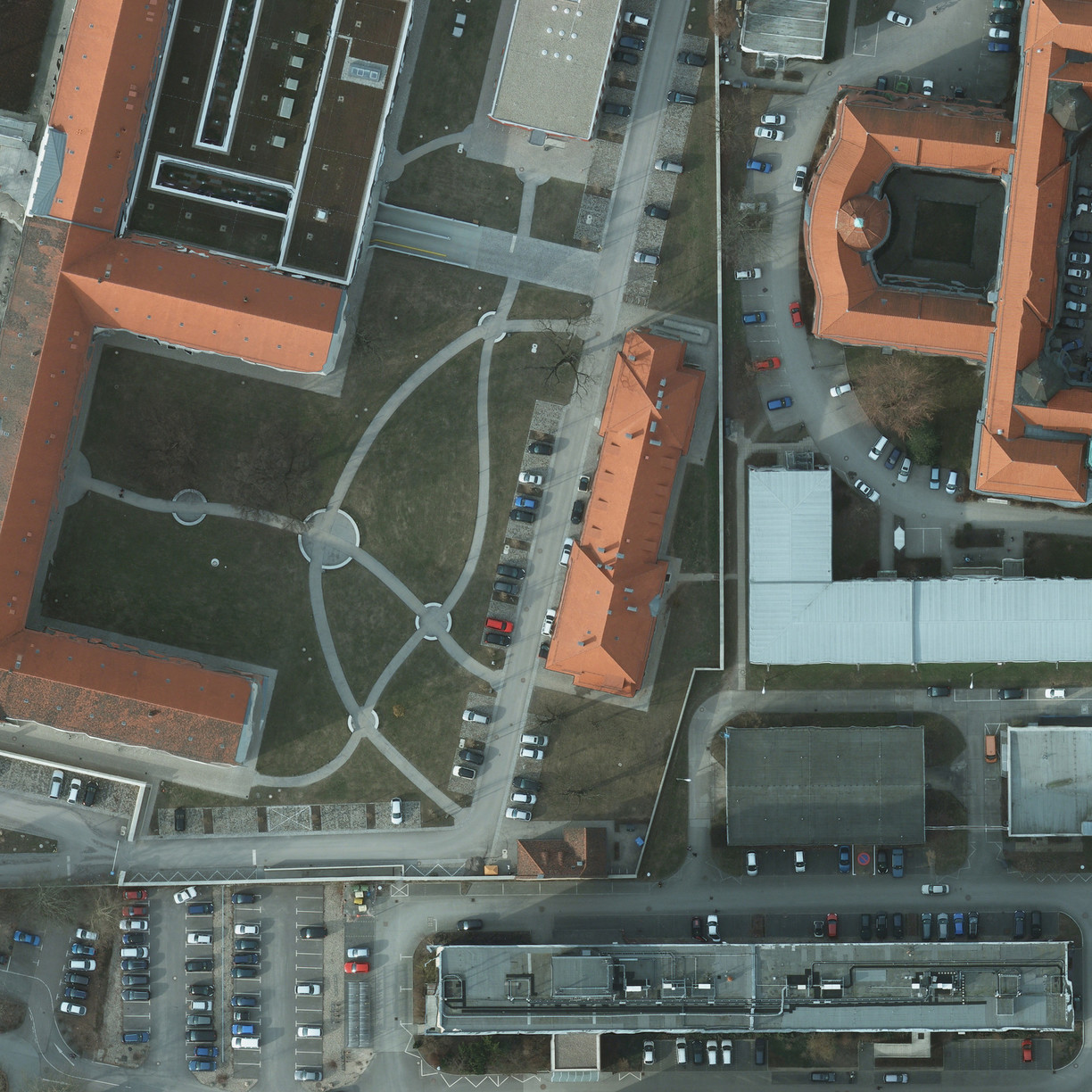}
        \caption*{Potsdam (RGB)}
        \end{subfigure}
        \begin{subfigure}{0.49\textwidth}
        \includegraphics[width=\textwidth]{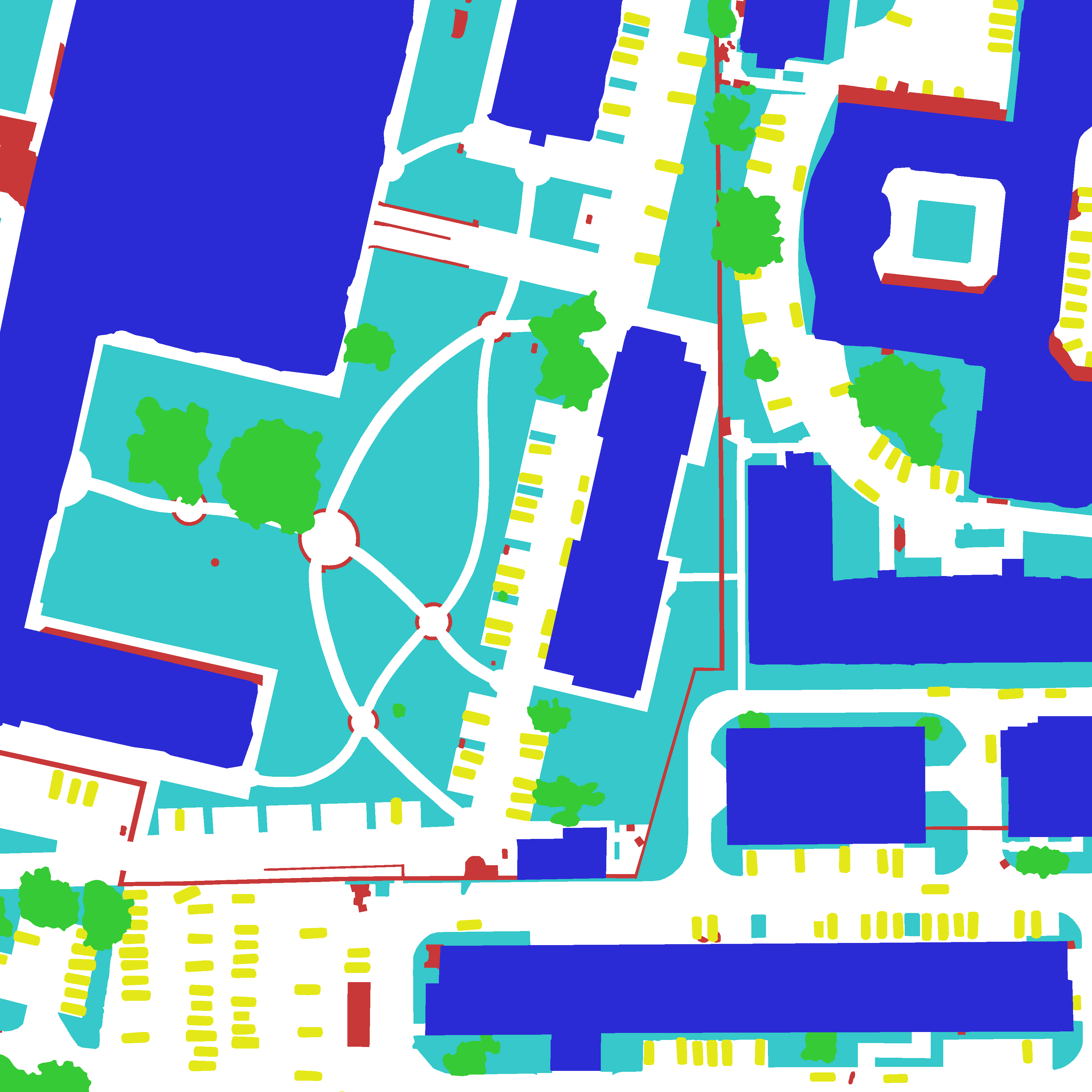}
        \caption*{Potsdam (GT)}
        \end{subfigure}
        \caption{ISPRS Potsdam dataset}
    	\label{fig:dataset_potsdam}
    \end{subfigure}
    \begin{subfigure}{0.49\textwidth}
    \begin{subfigure}{0.49\textwidth}
        \includegraphics[width=\textwidth]{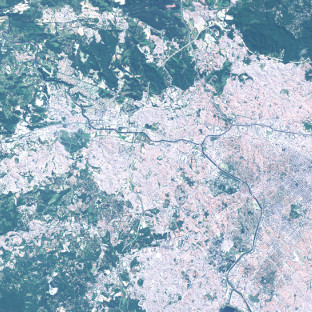}
        \caption*{DFC 2017 (false RGB)}
        \end{subfigure}
        \begin{subfigure}{0.49\textwidth}
        \includegraphics[width=\textwidth]{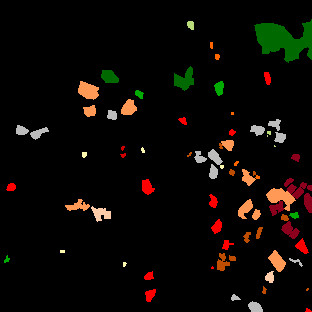}
        \caption*{DFC 2017 (GT)}
        \end{subfigure}
        \caption{DFC 2017 dataset (extract from São Paulo)}
    	\label{fig:dataset_dfc}
    \end{subfigure}
    \caption{Extract of the ISPRS Potsdam and DFC2017 datasets}
    \label{fig:datasets}
\end{figure*}

\paragraph{ISPRS Potsdam}

The ISPRS Potsdam Semantic Labeling dataset~\cite{rottensteiner_isprs_2012}\footnote{\url{http://www2.isprs.org/commissions/comm3/wg4/semantic-labeling.html}} is comprised of 38 ortho-rectified aerial IRRGB images ($6000 \times 6000$ px) at 5cm spatial resolution, taken over the city of Potsdam (Germany). A comprehensive pixel-level ground truth is provided for 24 tiles with annotations for the following classes: ``impervious surface'', ``building'', ``low vegetation'', ``tree'', ``car'' and ``clutter'', as illustrated in~\cref{fig:dataset_potsdam}. As the tiles are geo-referenced, we download and generate the corresponding OpenStreetMap rasters with the footprints for roads, buildings, vegetation and water bodies using Maperitive\footnote{\url{http://maperitive.net/}}. Results on this dataset are cross-validated on a 3-fold train/test split (18 tiles for training, 6 tiles for testing).

\paragraph{Data Fusion Contest 2017}
The Data Fusion Contest (DFC) 2017 dataset\footnote{\url{http://www.grss-ieee.org/community/technical-committees/data-fusion/data-fusion-contest/}} or ``grss\_dfc\_2017''~\cite{grss_2017_2017} is comprised of various acquisitions over 8 different cities: Berlin, Hong Kong, Paris, Rome and Sao Paulo for training, and Amsterdam, Chicago, Madrid an Xi'an for testing. It includes multispectral Sentinel-2 and Landsat data, OpenStreetMap rasters for the roads, vegetation, land use types, buildings and water bodies and a sparse ground truth containing annotations for several Local Climate Zones (LCZ), as illustrated in~\cref{fig:dataset_dfc}. The LCZ define urban or rural areas such as ``sparsely built urban area'', ``water body'', ``dense trees'' and so on, using the taxonomy from~\cite{stewart_local_2012}. The annotations cover only a part of the cities and are provided at 100m/pixel resolution. The goal is to generalize those annotations to the testing cities. In this work, we use the multispectral 20m/pixel resolution Sentinel-2 data and the OSM raster for roads, buildings, vegetation and water bodies. We preprocess the multispectral data by clipping the 2\% highest values. All 13 bands are kept and stacked as input to the neural network. As Sentinel-2 multispectral data includes bands at 10m/pixel, 20m/pixel and 60m/pixel resolutions, bands that have a resolution lower or higher than 20m/pixel are rescaled using bilinear interpolation. Results on this dataset are computed on the held-out testing set from the benchmark.

\subsection{Experimental setup}
We train our models on the ISPRS Potsdam dataset in an end-to-end fashion, following the guidelines from~\cite{audebert_semantic_2016}. We randomly extract $128\times128$ patches from the RGB and OSM training tiles on which we apply random flipping or mirroring as data augmentation. The optimization is performed with a batch size of 10 on the RGB tiles using a Stochastic Gradient Descent (SGD) with a learning rate of 0.01 divided by 10 every 2 epochs ($\simeq$ 30 000 iterations). SegNet's encoder for the RGB data is initialized using VGG-16~\cite{simonyan_very_2014} weights trained on ImageNet, while the decoder is randomly initialized using the MSRA~\cite{he_delving_2015} scheme. The learning rate for the encoder is set to half the learning rate for the decoder. During testing, each tile is processed by sliding a $128\times128$ window with a stride of 64 (\textit{i.e.} 50\% overlap). Multiple predictions for overlapping regions are averaged to smooth the map and reduce visible stitching on the patch borders. Training until convergence ($\simeq$ 150,000 iterations) takes around 20 hours on a NVIDIA K20c, while evaluating on the validation set takes less than 30 minutes.

On the DFC2017, we re-train SegNet from scratch and the weights are initialized using the MSRA scheme. As the input data has a resolution of 20m/pixel and the output LCZ are expected to be at 100m/pixel resolution, we use a smaller decoder by removing the last three convolutional blocks and the associated pooling layers. The resulting predictions have a resolution of 1:4 the input data and are interpolated to the 100m/pixel resolution. We train the network on randomly flipped $160\times160$ patches with a 50\% overlap. The patches are randomly selected but have to contain at least 5\% of annotated pixels. To avoid learning on non-labeled pixels from the sparse LCZ annotations, we ignore the undefined pixels in the ground truth during loss computation. The network is trained using the Adam~\cite{kingma_adam:_2014} optimizer with a learning rate of 0.01 with a batch size of 10. Training until convergence ($\simeq$ 60,000 iterations) takes around 6 hours on a NVIDIA Titan Pascal, while evaluating on the test set takes less than 5 minutes.

\begin{figure}
	\begin{subfigure}[t]{0.49\linewidth}
    	\includegraphics[width=\textwidth]{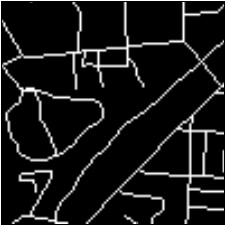}
        \caption{Binary representation.}
        \label{fig:osm_modeling_binary}
    \end{subfigure}
    \begin{subfigure}[t]{0.49\linewidth}
    	\includegraphics[width=\textwidth]{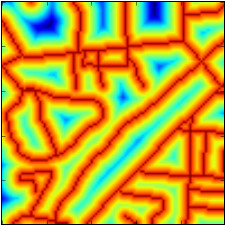}
        \caption{Signed distance transform.}
        \label{fig:osm_modeling_dist}
    \end{subfigure}
    \caption{Representations of the OSM layer for roads.}
    \label{fig:osm_modeling}
\end{figure}

OSM data modelization has to be carefully investigated. Most sensor data is continuous both in the numerical meaning but also in the spatial repartition. In many cases, if the original data is not continuous but sparse, well-chosen representations are used to get the continuity back, \eg the Digital Surface Model which is a continuous topology extracted from the sparse LiDAR point cloud. In our case, the OSM data is sparse and discrete like the labels. Therefore, it is dubious if the deep network will be able to handle all the information using such a representation. We compare two representations, illustrated in~\cref{fig:osm_modeling}:
\begin{itemize}
	\item Sparse tensor representation, which is discrete. For each raster, we have an associated channel in the tensor which is a binary map denoting the presence of the raster class in the specified pixel (cf.~\cref{fig:osm_modeling_binary}).
    \item Signed distance transform (SDT) representation, which is continuous. We generate for each raster the associated channel corresponding to the distance transform $d$, with $d > 0$ if the pixel is inside the class and $d < 0$ if not (cf.~\cref{fig:osm_modeling_dist}, with a color range from blue to red).
\end{itemize}

The signed distance transform was used in~\cite{yuan_automatic_2016} for building extraction in remote sensing data. The continuous representation helped densifying the sparse building footprints that were extracted from a public GIS database and successfully improved the classification accuracy.

\subsection{Results}

\begin{table*}[t]
	\caption{Test results on the ISPRS Potsdam dataset (pixel-wise overall accuracy and F1 score per class).}
    \label{table:potsdam_results}
	\begin{tabularx}{\textwidth}{c Y c c c c c c c}
    \toprule
    OSM & Method & imp. surfaces & buildings & low veg. & trees & cars & Overall\\
    \midrule
    Binary & OSMNet & 54.8 & 90.0 & 51.5 & 0.0 & 0.0 & 60.3\\
    $\emptyset$ &  SegNet RGB &	93.0		&	92.9		&		85.0		&	85.1	&	95.1	&	89.7\\ % Total accuracy : 89.6937874707% F1Score : imp_surfaces: 0.929809394067 building: 0.929299015161 low_vegetation: 0.849748586476 tree: 0.850844869105 car: 0.951459796967
    \midrule
    \multirow{2}{*}{Binary} &  Residual Correction RGB+OSM &	93.9	&	92.8		&		85.1		&	\textbf{85.2} &	95.8	&	90.6\\ % Total accuracy : 90.6055047156% F1Score : imp_surfaces: 0.939006408423 building: 0.927981495848 low_vegetation: 0.851468982357 tree: 0.851769654488 car: 0.95752065222
    & FuseNet RGB+OSM &	\textbf{95.3}	&	\textbf{95.9}		&		86.3	&	85.1	&	\textbf{96.8}	&	\textbf{92.3}\\% Total accuracy : 92.2906797436 imp_surfaces: 0.952975418574 building: 0.95940900256 low_vegetation: 0.863144254976 tree: 0.851434162082 car: 0.968110673807
    \midrule
    \multirow{2}{*}{SDT} & Residual Correction RGB+OSM &	93.8	&	92.7		&		85.2		&	84.8 &	95.9	&	90.5\\ % 90.5032708831% F1Score : imp_surfaces: 0.937538952263 building: 0.926691457575 low_vegetation: 0.852227032919 tree: 0.84842283045 car: 0.959002135002
    & FuseNet RGB+OSM &	95.2	&	\textbf{95.9}		&		\textbf{86.4}		&	85.0	&	96.5	&	\textbf{92.3}\\%92.2509199168 F1Score : imp_surfaces: 0.952434690995 building: 0.958692560397 low_vegetation: 0.864086862072 tree: 0.849330946994 car: 0.965473037252
    \bottomrule
    \end{tabularx}
\end{table*}

\begin{table*}[t]
	\caption{Test results on the DFC2017 dataset (pixel-wise accuracies)}
    \label{table:dfc_results}
    \setlength{\tabcolsep}{2pt}
	\begin{tabularx}{\textwidth}{c c c c c c c c c c c c c c c c c c c}
    \toprule
    \multirow{3}{*}{LCZ} & \multicolumn{10}{c}{Urban} & \multicolumn{7}{c}{Rural}\\
    & \multicolumn{3}{Y}{Compact} & \multicolumn{3}{Y}{Open} & \multicolumn{4}{Y}{Misc. buildings} & \multicolumn{2}{c}{Trees} & \multicolumn{2}{c}{Vegetation} & \multicolumn{3}{c}{Soils and water}\\
    & 1 & 2 & 3 & 4 & 5 & 6 & 7 & 8 & 9 & 10 & 11 & 12 & 13 & 14 & 15 & 16 & 17 &	OA\\
    \midrule
    SegNet multispectral &	\textbf{34.7} & 25.4 & 8.6 & \textbf{19.7} & \textbf{14.6} & 17.5 &	0.0 & \textbf{62.3} & 0.0 & \textbf{1.0} & 66.9 & \textbf{4.3} & \textbf{13.1} & 62.5 & 0.0 & 0.0 & \textbf{89.2} & 41.7\\
    FuseNet multispectral + OSM & 34.3 & \textbf{39.1} & \textbf{26.0} & 16.7 & 6.2 & \textbf{37.1} & 0.0 & 45.2 & \textbf{9.2} & 0.0 & \textbf{83.4} & 1.8 & 0.0 & \textbf{80.2} & \textbf{1.4} & 0.0 & 87.3 & \textbf{46.5}\\
    \bottomrule
    \end{tabularx}
\end{table*}

\paragraph{OSM data representation}
As can be seen in~\cref{table:potsdam_results}, learning and testing on the signed distance transform (SDT) representation for the OpenStreetMap layers performs slightly worse than its binary counterpart. This might seem counter-intuitive, as the distance transform contains a denser information. However, we suggest that this information might be too diffuse and that the model loses the sharp boundaries and clear transitions of the binary raster on some parts of the dataset. Yet, the difference between the two representations does not impact strongly the final accuracy.

\paragraph{ISPRS Potsdam}
We report in~\cref{table:potsdam_results} the results of our methods on our validation set of the ISPRS Potsdam dataset. In accordance with the dataset guidelines, we compare our predictions with the ground-truth where the borders have been eroded by a disk of radius 3 pixels. We report the overall pixel-wise accuracy and the F1 score for each class, computed as follows:
\begin{equation}
F1_{i} = 2~\frac{precision_{i} \times recall_{i}}{precision_{i} + recall_{i}}~,
\end{equation}
\begin{equation}
recall_i = \frac{tp_i}{C_i},~ precision_i = \frac{tp_i}{P_i}~,
\end{equation}
where $tp_i$ the number of true positives for class $i$, $C_i$ the number of pixels belonging to class $i$, and $P_i$ the number of pixels attributed to class $i$ by the model.

\begin{figure*}
\begin{subfigure}{0.33\linewidth}
	\includegraphics[width=\textwidth]{potsdam_rgb_4_12}
    \caption{RGB input}
\end{subfigure}
\hfill
\begin{subfigure}{0.33\linewidth}
    \includegraphics[width=\textwidth]{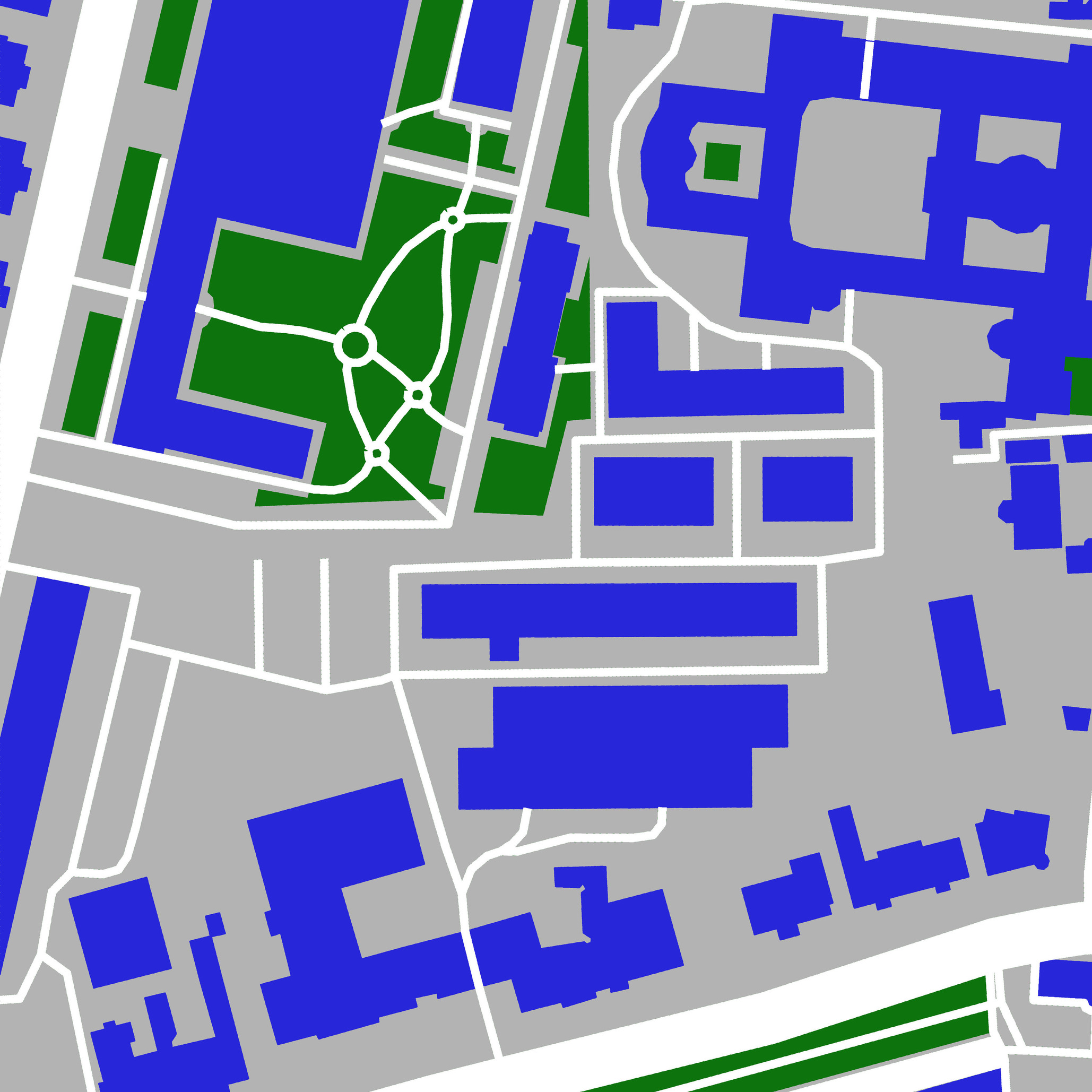}
    \caption{OSM input}
\end{subfigure}
\hfill
\begin{subfigure}{0.33\linewidth}
	\includegraphics[width=\textwidth]{potsdam_gt_4_12}
    \caption{Target ground truth}
\end{subfigure}
\begin{subfigure}{0.49\linewidth}
	\includegraphics[width=\textwidth]{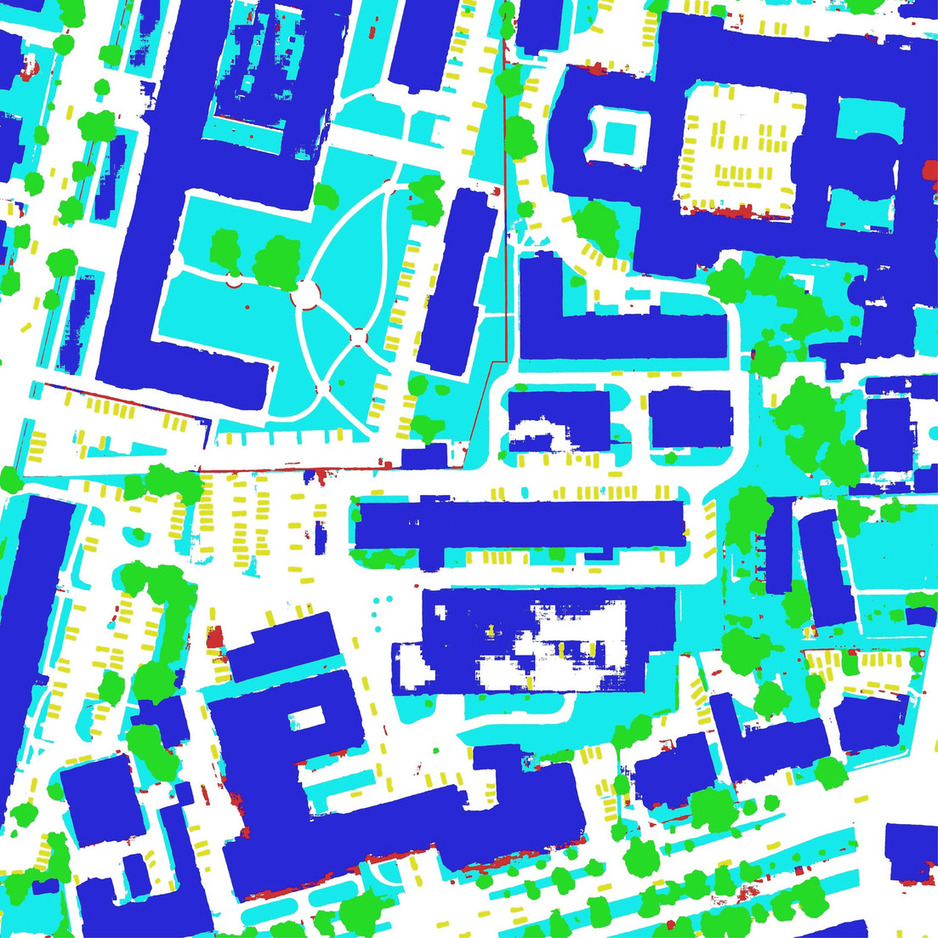}
    \caption{SegNet (RGB)}
\end{subfigure}
\hfill
\begin{subfigure}{0.49\linewidth}
	\includegraphics[width=\textwidth]{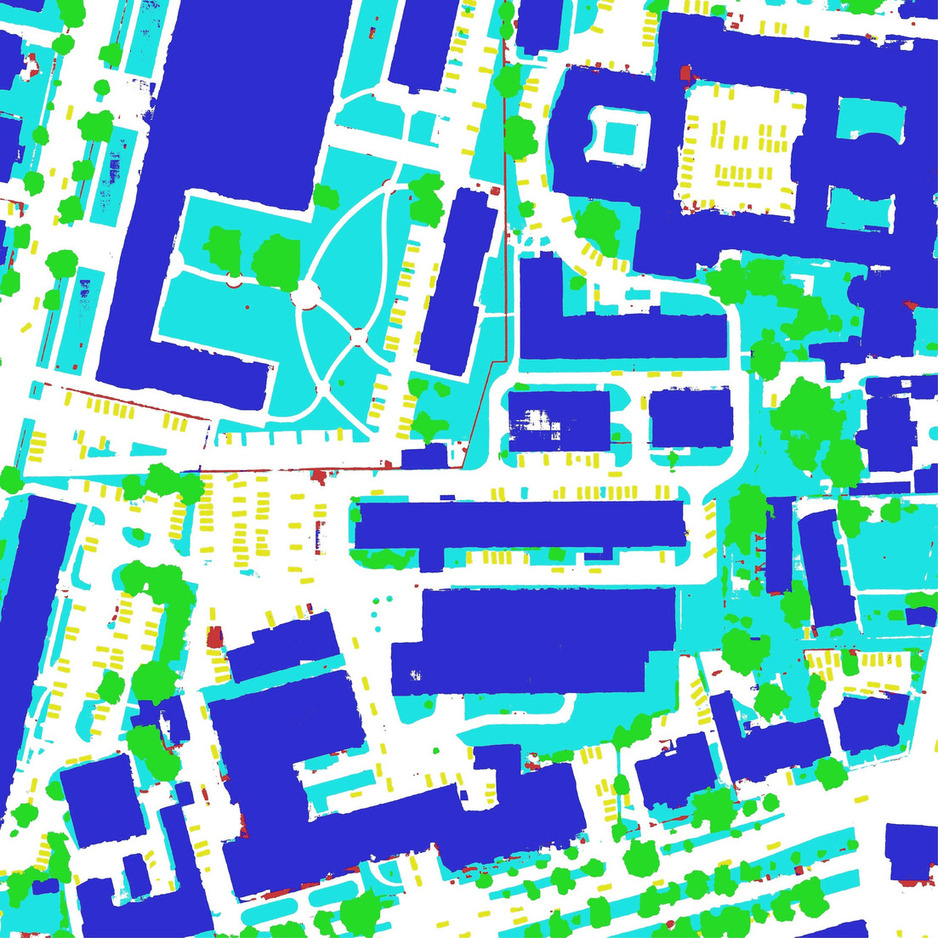}
    \caption{FuseNet (OSM+RGB)}
\end{subfigure}
\caption{Excerpt from the classification results on Potsdam}
\label{fig:potsdam_qualitative}
\end{figure*}

\begin{figure*}
\begin{subfigure}{0.33\linewidth}
	\includegraphics[width=\textwidth]{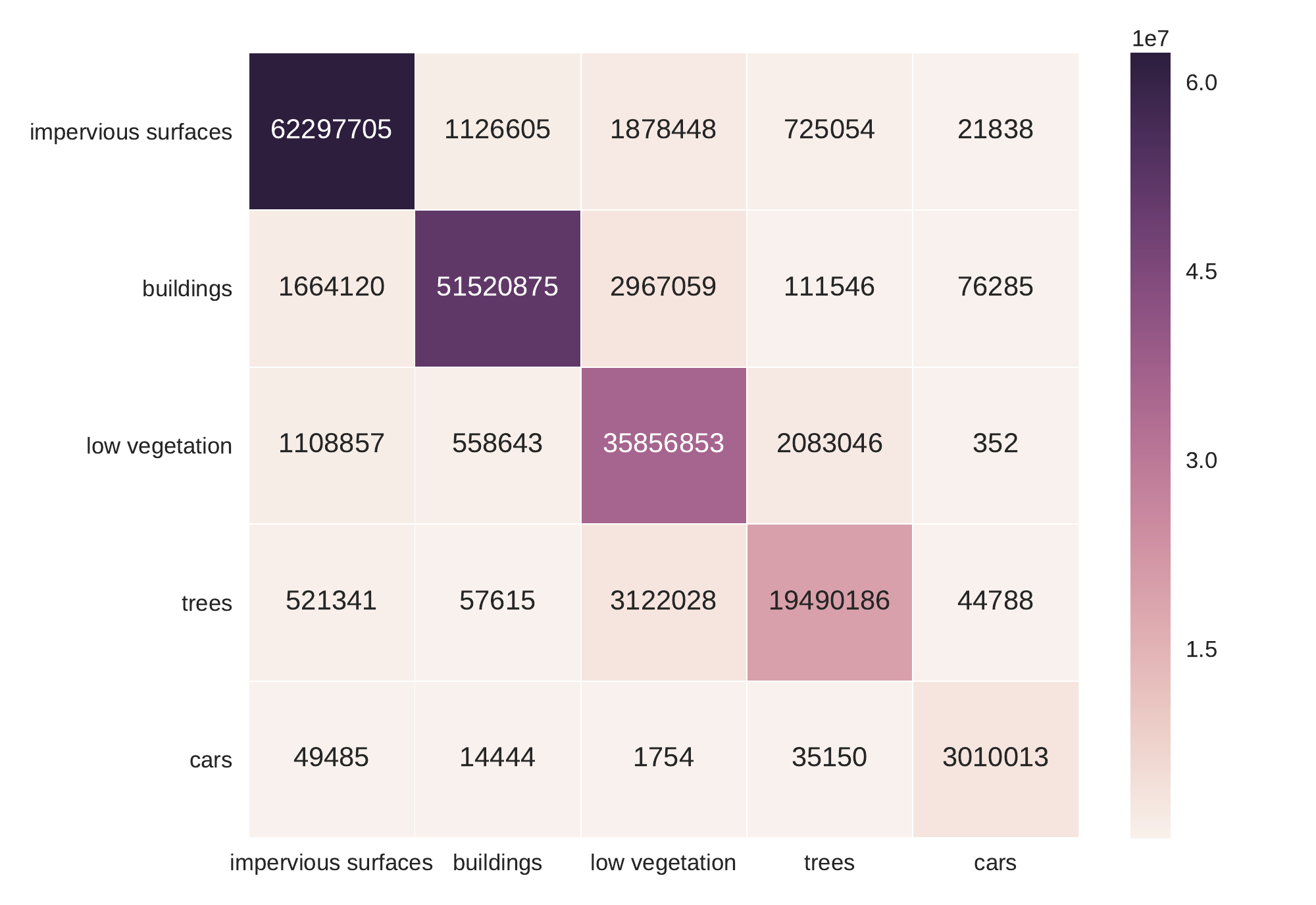}
    \caption{SegNet RGB}
\end{subfigure}
\begin{subfigure}{0.33\linewidth}
	\includegraphics[width=\textwidth]{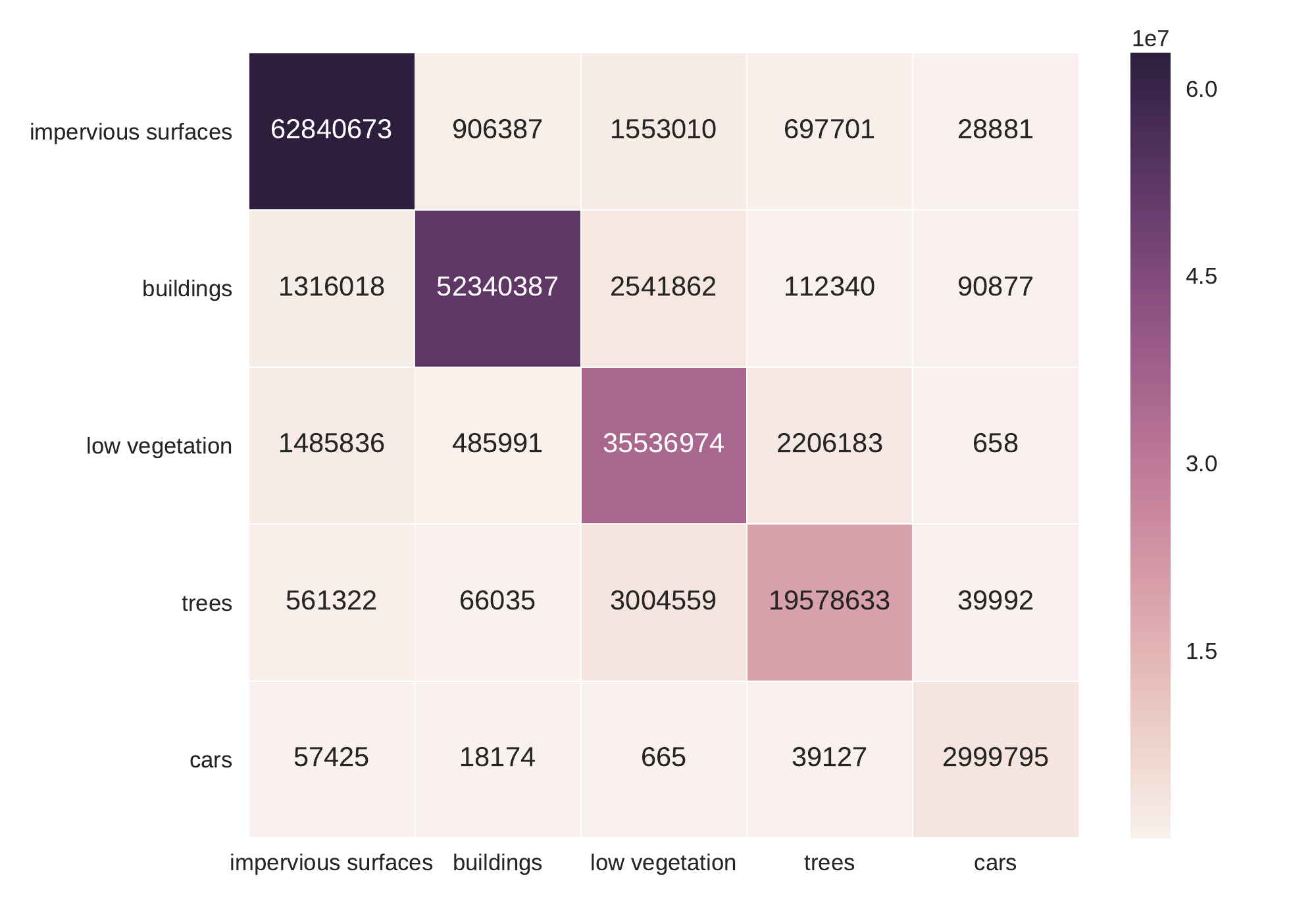}
    \caption{Residual Correction OSM+RGB}
\end{subfigure}
\begin{subfigure}{0.33\linewidth}
	\includegraphics[width=\textwidth]{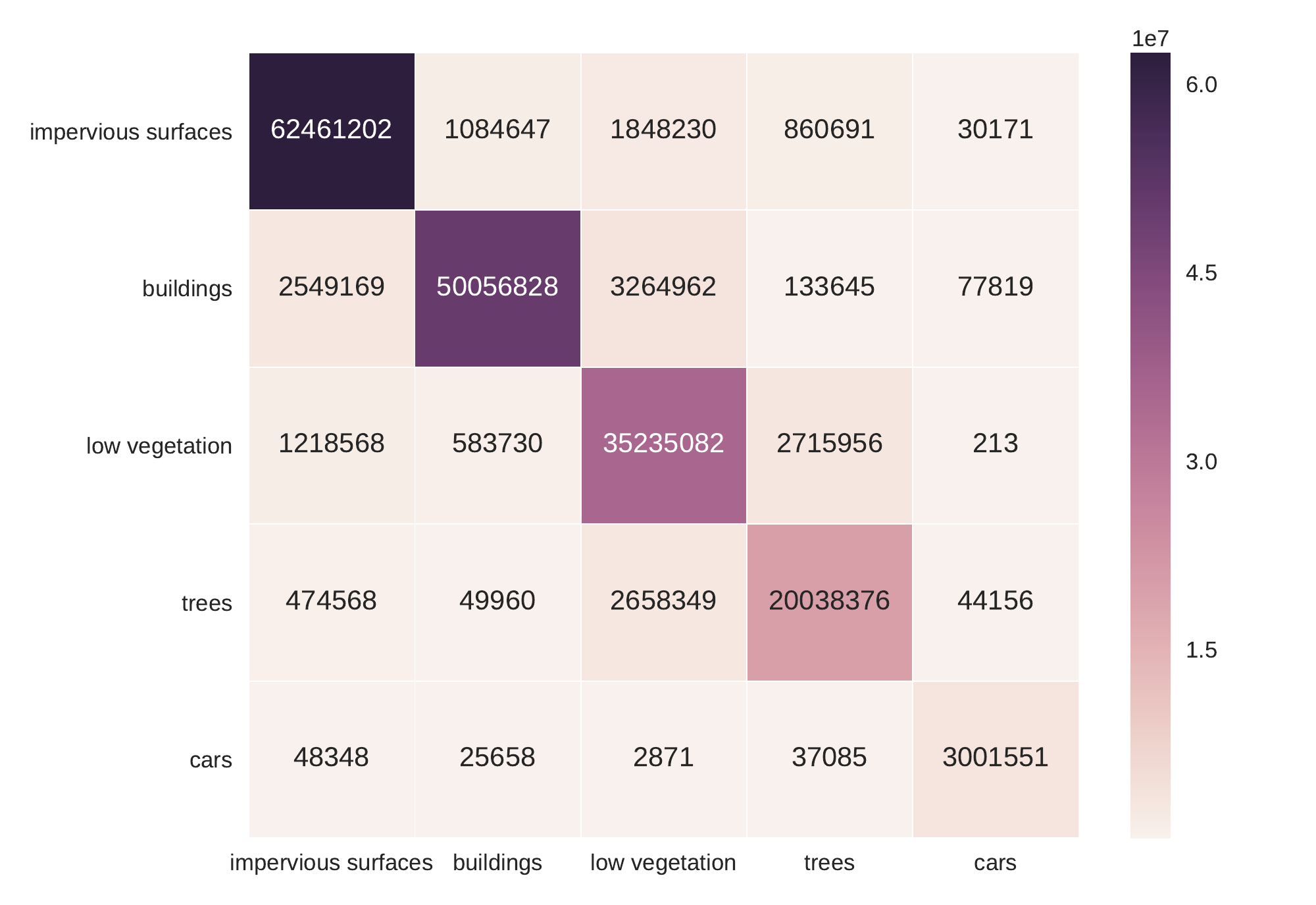}
    \caption{FuseNet OSM+RGB}
\end{subfigure}
\caption{Confusion matrices on Potsdam using the different methods.}
\label{fig:cm_potsdam}
\end{figure*}

As could be expected, the inclusion of OSM data improves the semantic labeling performance, with significant improvements on ``road'' and ``building'' classes. This is not surprising considering that those classes already have a representation in the OSM layers which can help disambiguating predictions coming from the optical source. This is illustrated qualitatively in~\cref{fig:potsdam_qualitative} and quantitatively in~\cref{fig:cm_potsdam}. Moreover, OSM data accelerates the training process as it allows the main network to focus on the harder parts of the image. Indeed, OSM data already covers the majority of the roads and the buildings, therefore simplifying the inference of the ``impervious surface'' and ``building'' classes. OSM data also helps discriminating between buildings and roads that have similar textures.

\begin{figure*}
\begin{tabularx}{\textwidth}{c Y Y Y Y Y Y Y}
RGB only & 
\includegraphics[width=0.1\textwidth]{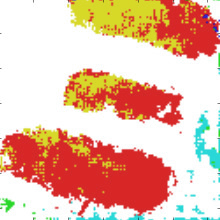} &
\includegraphics[width=0.1\textwidth]{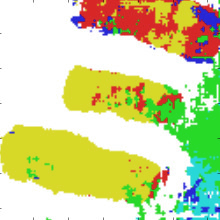} &
\includegraphics[width=0.1\textwidth]{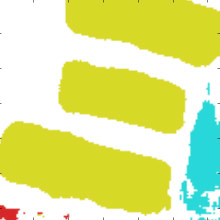} &
\includegraphics[width=0.1\textwidth]{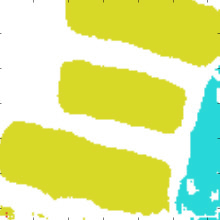} &
\includegraphics[width=0.1\textwidth]{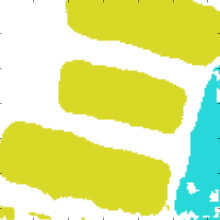} &
RGB &
\includegraphics[width=0.1\textwidth]{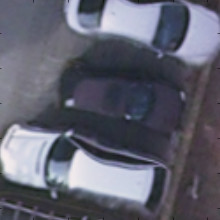} \\
RGB + OSM &
\includegraphics[width=0.1\textwidth]{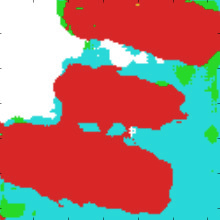} &
\includegraphics[width=0.1\textwidth]{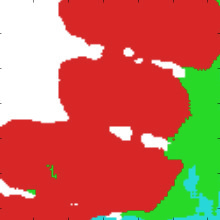} &
\includegraphics[width=0.1\textwidth]{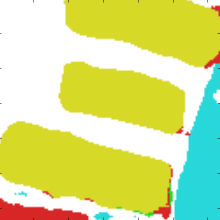} &
\includegraphics[width=0.1\textwidth]{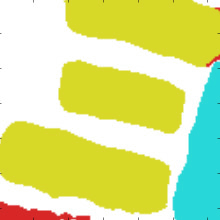} &
\includegraphics[width=0.1\textwidth]{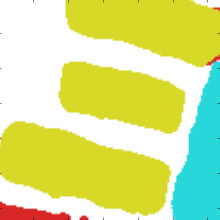} &
GT &
\includegraphics[width=0.1\textwidth]{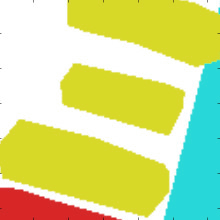} \\
iteration & 10,000 & 20,000 & 50,000 & 80,000 & 120,000\\
\end{tabularx}
\caption{Evolution of the predictions coming from SegNet using RGB only vs. RGB + OSM. Integrating OSM data makes the output more visually coherent, even in the early learning stages.\\
\small{Legend: white: impervious surfaces, \textcolor{Blue}{blue}: buildings, \textcolor{Cerulean}{cyan}: low vegetation, \textcolor{OliveGreen}{green}: trees, \textcolor{Dandelion}{yellow}: vehicles, \textcolor{BrickRed}{red}: clutter}, black: undefined.}
\label{fig:training_evolution}
\end{figure*}

An interesting side effect of the integration of the OSM data into the learning process is the significant speedup in convergence time that can be observed. Indeed, on the same dataset, the coarse-to-fine model converges approximately 25\% faster to the same overall accuracy compared to the classical RGB SegNet, \textit{i.e.} the network requires 25\% less iterations to reach the same classification performance. Moreover, this accuracy is reached with a mean loss of $0.45$ for the latter, while the former has a mean loss of only $0.39$, which indicates a better generalization capability. This is similar to the findings from~\cite{he_deep_2016} on residual learning. Finally, the inclusion of the OSM data helps regularizing spatially the predictions when the network is still in early training.~\cref{fig:training_evolution} illustrates how the same patch, classified at several stages in the training, is visually better represented using both OSM and RGB sources compared to the RGB image only.

\paragraph{DFC2017}
\begin{figure}
\begin{subfigure}{0.49\linewidth}
	\includegraphics[width=\textwidth]{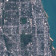}
    \caption{RGB (composite)}
\end{subfigure}
\begin{subfigure}{0.49\linewidth}
	\includegraphics[width=\textwidth]{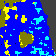}
    \caption{Prediction (FuseNet)}
\end{subfigure}
\caption{Partial results on the city of Chicago (DFC2017)\\
\small{Legend: \textcolor{Cerulean}{cyan}: compact high rise, \textcolor{Blue}{blue}: compact mid-rise, \textcolor{Dandelion}{yellow}: open high rise, \textcolor{Brown}{brown}: dense trees, \textcolor{Gray}{grey}: water}}
\label{fig:dfc_chicago}
\end{figure}

We report in~\cref{table:dfc_results} the detailed results of our methods on the DFC2017 dataset. The evaluations metrics are the overall accuracy (OA) and the pixel-wise accuracies for each class. The integration of the OSM layers is mostly useful for two types of classes: the high density urban areas and the dense vegetation. Indeed, classes 1, 2 and 3 denote different sorts of compact urban areas. The buildings footprints from OSM help detecting those classes and estimating the local building density. For the vegetation, the relevant OSM annotations are in the natural terrain layer. These annotations are mostly concentrated on forests and large meadows, which belong to the classes 11 (``dense trees'') and 14 (``low plants''). Consequently, those classes obtain a significant increase in classification accuracy when fusing OSM and multispectral data. A qualitative example on the city of Chicago is shown in~\cref{fig:dfc_chicago}.

\section{Discussion}
The methods presented in this work allow us to improve semantic labeling of aerial and satellite multispectral images thanks to the integration of several OSM rasters, notably the roads, buildings and vegetation land use. However, OpenStreetMap data is much more exhaustive than such layers and also contains specific information (\eg swamps, agriculture fields, industrial areas, different categories of roads\dots). However, if all information are stacked by using one map per layer of interest, the OSM memory footprint would become huge very quickly, especially considering that OSM provides vector information that can be rasterized at any spatial resolution. In our case, we rasterize the OSM layers to the same resolution as our input image, which can be very high for airborne acquisitions. Moreover, we have not addressed here the question of the subclassification, while this is definitely a source of future improvement. Indeed, thanks to OSM data, we can know that some specific buildings have a particular type, \eg a building can be a church, a grocery store or a house. Point annotations, such as parking lots, are also dismissed but could provide meaningful insights about the semantics of the area.
Furthermore, we underline that even though the OSM layers that we used were more recent (2 years) than to the optical data, there were few enough disagreements so that the models were robust to those conflicts. Yet, data fusion should be done carefully if the sources do not represent the same underlying reality. In the case of the OSM data, this could be worked around by extracting the layers from the OSM archives if the optical data is not recent enough. Finally, mapping style and coverage can vary a lot based on the observed regions. For example, urban areas in developed countries are thoroughly mapped, whereas annotations are very scarce in rural areas in developing countries. This enforces the need for the model to be robust to errors and missing OSM input data for very large scale mapping.

\section{Conclusion}
In this work, we showed how to integrate ancillary GIS data in deep learning-based semantic labeling. We presented two methods: one for coarse-to-fine segmentation, using deep learning on RGB data to refine OpenStreetMap semantic maps, and one for data fusion to merge multispectral data and OSM rasters to predict local climate zones. We validated our methods on two public datasets: the ISPRS Potsdam 2D Semantic Labeling Challenge and the Data Fusion Contest 2017. We increase our semantic labeling overall accuracy by 2.5\% on the former and by nearly 5\% on the latter by integrating OpenStreetMap data in the learning process. Moreover, on the ISPRS Potsdam dataset, using OSM layers in a residual correction fashion accelerates the model convergence by 25\%. Our findings show that GIS sparse data can be leveraged successfully for semantic labeling on those two use cases, as it improves significantly the classification accuracy of the models. We think that using crowdsourced and open GIS data is an exciting topic of research, and this work provides new insights on how to use this data to improve and accelerate learning based on traditional sensors.

\section*{Acknowledgements}
The Vaihingen dataset was provided by the German Society for Photogrammetry, Remote Sensing and Geoinformation (DGPF) \cite{cramer_dgpf_2010}: \url{http://www.ifp.uni-stuttgart.de/dgpf/DKEP-Allg.html}. The authors thank the ISPRS for making the Vaihingen and Potsdam datasets available and organizing the semantic labeling challenge. The authors would like to thank the WUDAPT (\url{http://www.wudapt.org/}) and GeoWIKI (\url{http://geo-wiki.org/}) initiatives and the IEEE GRSS Image Analysis and Data Fusion Technical Committee.
Nicolas Audebert's work is supported by the Total-ONERA research project NAOMI.

{\small
\bibliographystyle{ieee}
\bibliography{CVPREarthvision}
}

\end{document}